\pdfoutput=1

\documentclass[11pt]{article}

\usepackage[]{acl}

\usepackage{times}
\usepackage{latexsym}

\usepackage[T1]{fontenc}

\usepackage[utf8]{inputenc}

\usepackage{microtype}

\usepackage{inconsolata}

\usepackage{graphicx}
\usepackage{tabularx}

\usepackage[normalem]{ulem}
\useunder{\uline}{\ul}{}
\usepackage{afterpage}
\usepackage{todonotes}
\usepackage{multirow}
\usepackage{arydshln}
\usepackage{tablefootnote}
\title{Planning for Success: Exploring LLM Long-term Planning Capabilities in Table Understanding}

\author{Thi-Nhung Nguyen$^1$, Hoang Ngo$^{2}$, Dinh Phung$^1$, Thuy-Trang Vu$^1$, Dat Quoc Nguyen$^2$ \\
  $^1$Monash University, $^2$Qualcomm AI Research\thanks{Qualcomm Vietnam Company Limited. Qualcomm AI Research is an initiative of Qualcomm Technologies, Inc.} \\
  \texttt{\{nhung.thinguyen,dinh.phung,trang.vu1\}@monash.edu} \\
  \texttt{\{hoangngo,datnq\}@qti.qualcomm.com}
}
\begin{document}
\maketitle
\begin{abstract}
 Table understanding is key to addressing challenging downstream tasks such as table-based question answering and fact verification. Recent works have focused on leveraging Chain-of-Thought and question decomposition to solve complex questions requiring multiple operations on tables. However, these methods often suffer from a lack of explicit long-term planning and weak inter-step connections, leading to miss constraints within questions. In this paper, we propose leveraging the long-term planning capabilities of large language models (LLMs) to enhance table understanding. Our approach enables the execution of a long-term plan, where the steps are tightly interconnected and serve the ultimate goal, an aspect that methods based on Chain-of-Thought and question decomposition lack. In addition, our method effectively minimizes the inclusion of unnecessary details in the process of solving the next short-term goals, a limitation of methods based on Chain-of-Thought. Extensive experiments demonstrate that our method outperforms strong baselines and achieves state-of-the-art performance on WikiTableQuestions and TabFact datasets.
\end{abstract}

\section{Introduction}

\begin{figure}[!t]
    \centering
\includegraphics[width=\columnwidth]{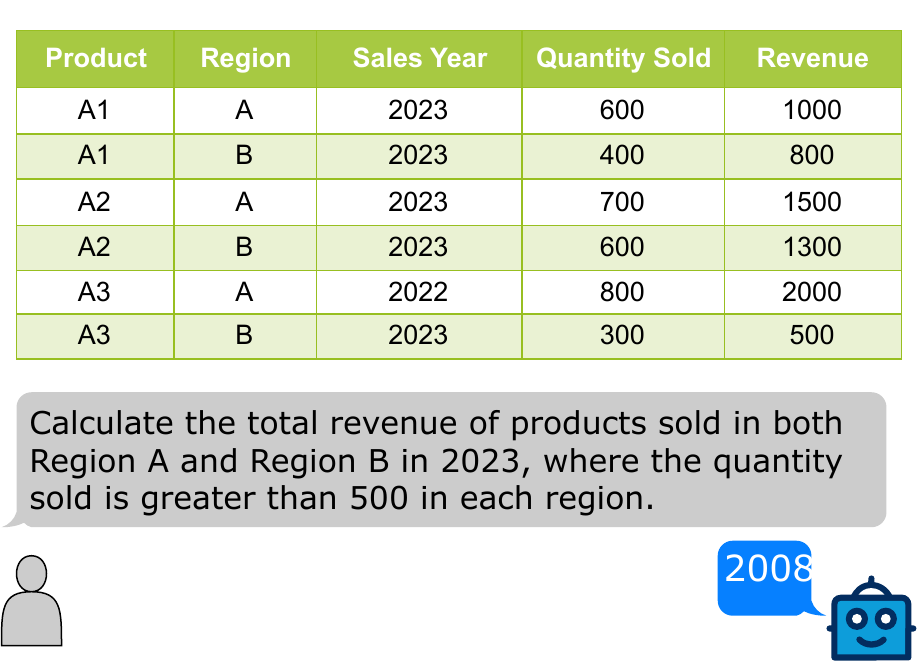}
    \caption{A question-answering example over a table.}
    \label{fig:problem}
\end{figure}

Table understanding is key to addressing challenging downstream tasks involving tables, one of the most prevalent forms of semi-structured data in real-world scenarios, such as table question answering \cite{wang2023surveytableandtexthybridqaconcepts,lin-etal-2023-inner} and fact verification \cite{Chen2020TabFact}. The primary goal is to accurately extract relevant information from tables to provide precise answers to user questions. To better understand the problem consider the example in Table~\ref{fig:problem}.

Early works focus on fine-tuning BERT to encode tables \cite{herzig2020tapas, Chen2020TabFact}. The key idea is to leverage specialized embedding layers or attention mechanisms to encode table cells or segments effectively, enabling models to understand the structure of tables. Another direction revolves around the synthesis of SQL query-response pairs to pre-train an encoder-decoder model as a neural SQL executor \cite{eisenschlos-etal-2020-understanding, liu2022tapex, jiang-etal-2022-omnitab}. With the advent of large language models (LLMs), recent works have explored instruction fine-tuning of LLMs with tabular data to create generalist models capable of handling a variety of table-based tasks \cite{zhang-etal-2024-tablellama}, showing improved performance over flagship closed-source LLMs such as \texttt{GPT-3.5-turbo} and \texttt{GPT-4} \cite{openai2024gpt4technicalreport}. 

\begin{figure}[!t]
    \centering
\includegraphics[width=\columnwidth]{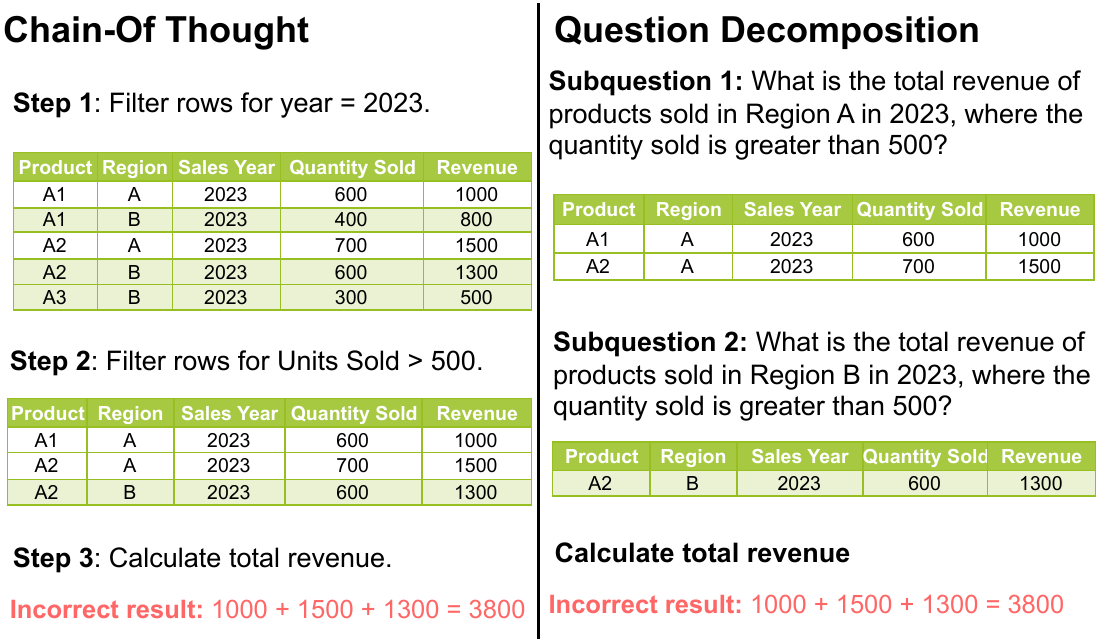}
    \caption{An illustration showing how CoT-based methods and question decomposition-based methods miss the important inter-region condition in revenue calculation (corresponding to the table and question in Figure~\ref{fig:problem}).}
    \label{fig:prior_works}
\end{figure}

Leveraging the strong in-context learning performance of LLMs, recent works have increasingly focused on addressing table understanding through prompting. One common approach is to convert the question into executable languages, allowing the use of tools such as SQL or Python to access the information inside the table \cite{lin-etal-2023-inner, gemmell-dalton-2023-toolwriter, wang2024chainoftable, nahid-rafiei-2024-tabsqlify, liu-etal-2024-rethinking, kong2024opentab}. However, due to the constraints of the single-pass generation process, these methods often struggle with complex questions requiring multiple steps of table operations. To address this challenge, some state-of-the-art methods employ Chain-of-Thought (CoT) reasoning, which enables multi-step reasoning \cite{yao2023tree, chen2023program, wei2022chain, wang2024chainoftable}. Others rely on question decomposition, breaking down the question into sub-questions, solving them individually, and finally synthesizing a final answer \cite{kong2024opentab, patnaik2024cabinet, 10.1145/3539618.3591708}. 
However, both CoT-based methods and question decomposition-based methods suffer from a lack of explicit long-term planning and weak inter-step connections. This results in missing constraints within the question, leading to incorrect final answers. An illustration of this issue is shown in Figure~\ref{fig:prior_works}, where step 3 is not conditionally linked to the previous steps. 
In addition, in the case of CoT-based methods, the entire current chain is often utilized to generate the output for the subsequent step. This approach can result in LLMs forgetting critical details or generating hallucinations, as they process a substantial amount of information, including extraneous details, which may introduce unnecessary complexity and lead to errors \cite{jiang-etal-2022-omnitab, chen2023large}.

In this paper, we propose leveraging the long-term planning capabilities of LLMs to address these challenges. Unlike methods based on CoT and question decomposition, which lack explicit long-term planning, our method begins with the formulation of a long-term plan upon receiving a question. This plan outlines the necessary steps, called short-term goals, to progress systematically from the initial table to the final answer. The short-term goals can be either independent or interconnected, depending on the requirements of the question, ensuring that each serves the long-term goal. To handle each short-term goal effectively, we leverage a set of specialized experts, each dedicated to a specific task. These experts take responsibility for handling short-term goals relevant to their specialization, operating independently to resolve the goals within their localized scope. At this local level, each expert focuses solely on their assigned goal without being influenced by other parts of the long-term plan. The intermediate steps executed by the execution experts are single-pass. Once the short-term goal is completed, only the final results are updated within the long-term plan, minimizing the inclusion of unnecessary information in the process of solving the next short-term goals---a common issue in CoT-based methods.

Our contributions are summarized as follows:  \textbf{(I)} We propose leveraging the long-term planning capabilities of LLMs to enhance table understanding. \textbf{(II)} Our approach enables the execution of a long-term plan where the steps are tightly interconnected, all serving the ultimate goal---an aspect that methods based on Chain-of-Thought and question decomposition lack. \textbf{(III)} Our approach effectively minimizes the inclusion of unnecessary details in the process of solving the next short-term goals---a limitation of methods based on Chain-of-Thought. \textbf{(IV)} Comprehensive experiments demonstrate that our approach achieves state-of-the-art performance, outperforming existing strong baselines on standard benchmarks WikiTableQuestions and TabFact.

\begin{figure*}[!t]
    \centering
    \includegraphics[width=0.99\textwidth]{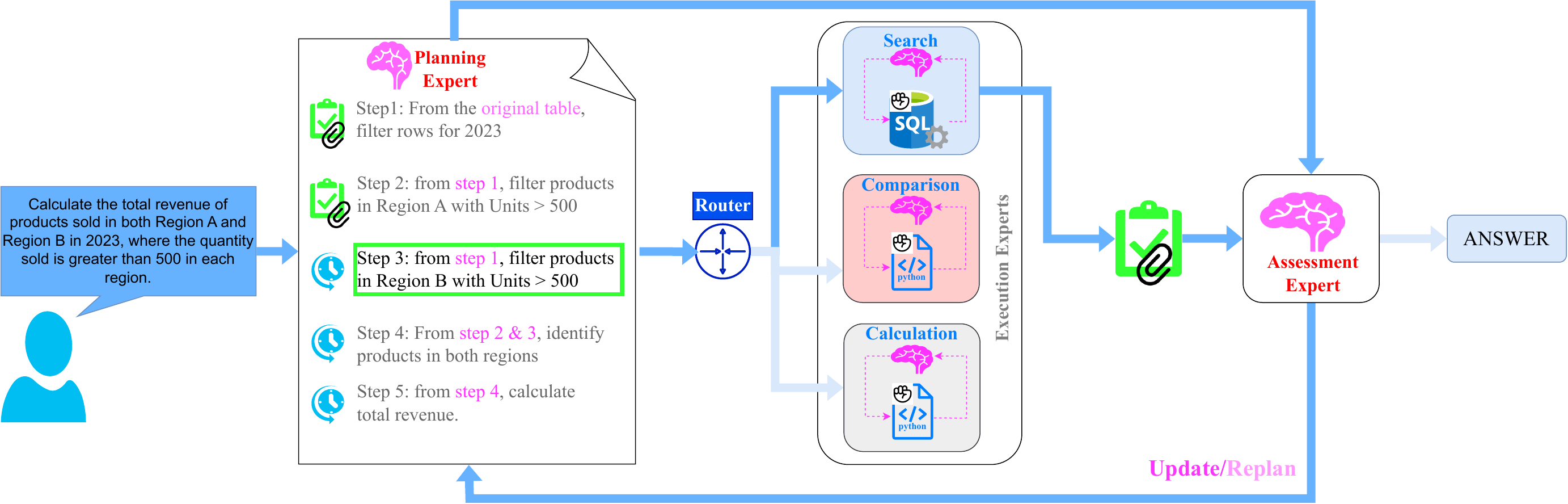}
    \caption{Overview of our proposed framework \textbf{PLANTA}.}
    \label{fig:method}
\end{figure*}

\section{Related Works}

Fine-tuning pre-trained BERT models \cite{kenton2019bert} were one the dominant approach for Table Understanding \cite{herzig2020tapas, Chen2020TabFact, liu2022ptab, deng2022turl, wang2021tuta, iida2021tabbie}. TaPas \cite{herzig2020tapas} leverage the mask language modeling approach proposed in BERT  to reconstruct certain cells in the table during training process. \citet{wang2021tuta} further enhance the performance by masking the entire columns in tables.
A different approach is to train an encoder-decoder model to transform questions into SQL queries and then answer these questions by executing the respective generated SQL queries \cite{eisenschlos-etal-2020-understanding, liu2022tapex, jiang-etal-2022-omnitab}. Recently, large language models (LLMs) have demonstrated excellent performance on a variety of tasks. Recent works have been shifting their focus to fine-tuning open-source LLMs to create models capable of handling a variety of table-based tasks. However, these methods require expensive labeled data and high training costs. This has led to the emergence of prompt-based approaches, which leverage the in-context learning capabilities of LLMs.

For prompt-based methods, some works propose concatenating task descriptions with the serialized table as a string and inputting them into an LLM to generate a text-based response \cite{marvin2023prompt, cheng2023binding, 10.1145/3616855.3635752}. Other works enhance the performance further by adding few-shot and curated examples to the prompt \cite{cheng2023binding, narayan2022can, chen2023large}. However, with reasoning only, LLMs often struggle to accurately retrieve all relevant data required within tables. Therefore, recent works increasingly incorporate external tools (e.g., Python and SQL) instead of relying solely on general text processing to effectively extract relevant data within tables \cite{chen2023program, gao2023pal, rajkumar2022evaluating, cheng2023binding, ni2023lever}. Despite this, due to the constraint of a single pass, this approach still struggles with complex questions where multiple operations need to be executed to produce an accurate answer.  
Recent state-of-the-art methods mitigate this limitation by employing chain-of-thought (CoT) reasoning or question decomposition \cite{chen2023program, zhao2023docmath, yang2024effective, zhou2023leasttomost, khot2023decomposed}.  Some works \cite{10.1145/3539618.3591708, cheng2023binding, liu-etal-2024-rethinking} further enhance the performance by self-consistency technique \cite{wang2023selfconsistency}, where a diverse set of reasoning paths is sampled from LLMs and the most consistent answer is selected to obtain the final answer. However, both CoT-based methods and question decomposition-based methods suffer from a lack of explicit long-term planning and weak inter-step connections. This results in constraints within the question being missed, leading to incorrect final answers. Furthermore, CoT-based methods often utilize the entire current chain to generate the output for the subsequent step. This approach can result in LLMs forgetting critical details or generating hallucinations, as they process a substantial amount of information, including extraneous details, which may introduce unnecessary complexity and lead to errors \cite{jiang-etal-2022-omnitab, chen2023large}.

\begin{figure*}[!t]
    \centering
    \includegraphics[width=\textwidth]{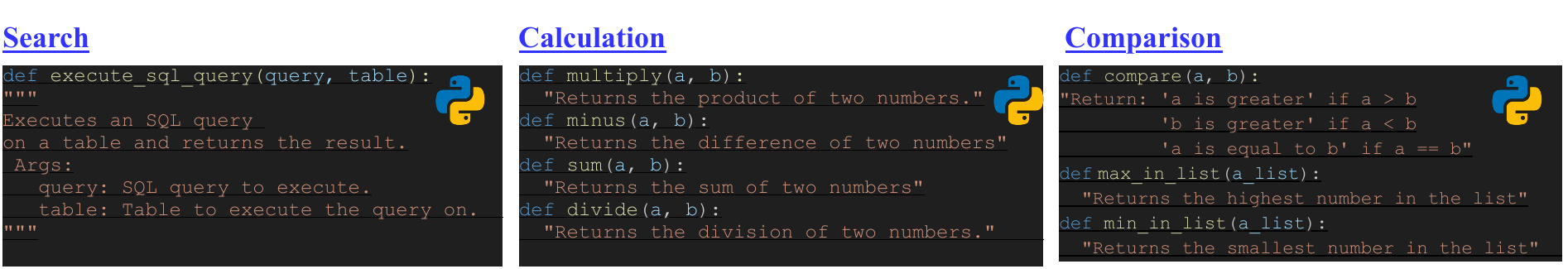}
    \caption{Predefined Python functions ("hands") assigned to the Search expert, Calculation expert, and Comparison expert in PLANTA, respectively.}
    \label{fig:hands}
\end{figure*}

\section{Our Approach}
We introduce a novel method, named \textbf{PLANTA}, which leverages the long-term \textbf{Plan}ning capabilities of Large Language Models to improve \textbf{Ta}ble Understanding. PLANTA is designed to tackle the challenge of generating accurate answers to table-based questions by extracting and reasoning over relevant information from the given tables.

Figure~\ref{fig:method} illustrates the architectural overview of PLANTA. First, upon receiving a table and a question, a \underline{Planning} expert comes up with a long-term plan outlining the necessary steps, called short-term goals, to transform the initial table into the desired answer to the user's question. Next, each short-term goal is routed to an appropriate \underline{Execution} expert by a \underline{Router}, which assigns short-term goals to experts based on their specialization via LLM prompting. These goals are then resolved locally, with only the final results passed to the following components of PLANTA, potentially updating the long-term plan. Meanwhile, intermediate steps executed by the \underline{Execution} experts are processed in a single pass. After each step, the updated long-term plan is evaluated by an \underline{Assessment} expert, who determines whether sufficient evidence has been gathered to answer the question or if modifications to the plan are necessary. If no adjustments are needed, the process continues. 
Below, we provide a detailed description of the architecture and roles of the experts within {PLANTA}. We first outline the common architecture shared by all experts in Subsection~\ref{common_architecture}, followed by an in-depth discussion of the differences in their architecture and their specific contributions in Subsection~\ref{task-specific_architecture}.

\subsection{Common Architecture}
\label{common_architecture}
In PLANTA, each expert consists of two main components: the "brain" and the "hands". Each \textit{brain} is specialized in a specific task and can independently reason to complete an assigned task. It is powered by an LLM, whose knowledge scope is encoded through prompting. The \textit{hands} are predefined tools, such as Python or SQL execution functions, tailored to the expert's specializations. These tools enable access to detailed data within tables and execute operations that LLMs may struggle with, such as calculations. They provide the brain with the necessary inputs for reasoning and determining the subsequent steps required to complete the task.

\subsection{Task-Specific Architecture}
\label{task-specific_architecture}

\paragraph{Planning expert:} Its role is to outline the necessary steps of short-term goals, structured as a task list, to transform the initial table into an accurate answer. Since this role focuses solely on planning without execution, the Planning expert's architecture comprises only the "brain". This brain is powered by an LLM specifically designed for the planning task, with a knowledge scope that includes the given table, the question, and the specializations of Execution experts (see our prompt for Planing in Appendix). For each step in the plan, dependencies on previous steps must be explicitly defined to enable the flexible reuse of variables from earlier steps. This approach minimizes the transfer of unnecessary information to subsequent steps while ensuring that all dependencies are correctly managed. For example, in Figure~\ref{fig:method}, step 3 depends only on the output of step 1. Therefore, step 3 can access only the output of step 1 that it depends on, without accessing the output of step 2.





\begin{figure}[!t]
    \centering
    \includegraphics[width=\columnwidth]{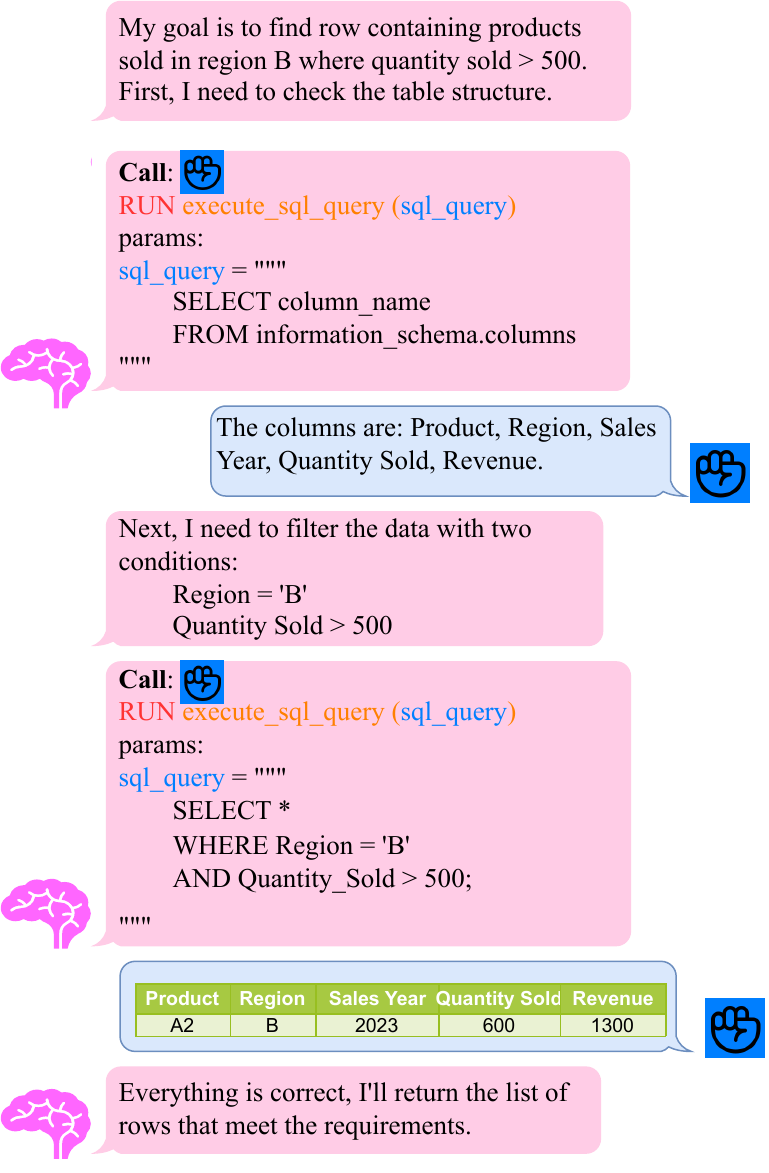}
    \caption{An example of how the Search expert addresses the 3rd short-term goal from Figure~\ref{fig:method}. Once the short-term goal is received, the Search expert performs reasoning step by step. The process includes understanding the question, analyzing the table structure, identifying the data that meets the goal's conditions, and providing the answer. When the expert needs to access data in the table, it automatically calls the predefined SQL execution function (see Figure~\ref{fig:hands}) to retrieve the necessary information. This is done by generating SQL queries as parameters for the function.}
    \label{fig:execution}
\end{figure}

\paragraph{Execution experts:} 
The task of the Execution experts is to resolve the short-term goals required by the Planning expert. These goals are assigned to appropriate Execution experts based on their specialization by the Router, which we perform using an LLM via prompting (see our prompt for the Router in Appendix).
Motivated by operations on tables, our PLANTA system is designed with three Execution experts: (1) Search expert, (2) Comparison expert, and (3) Calculation expert. In terms of their knowledge scope, they can only access the data and short-term goals provided by the Planning expert and reason with the support of predefined functions, which we call "hands". The predefined functions include SQL query execution for the Search expert; comparative and superlative comparison for the Comparison expert; and basic calculations such as addition, subtraction, multiplication, and division for the Calculation expert. See Figure \ref{fig:hands} for more details.

Unlike the initial question on tables, which must be addressed according to a pre-defined long-term plan, short-term goals are often simpler and more manageable. Therefore, we allow the experts to independently determine how to resolve assigned goals, such as utilizing the Chain-of-Thought or question decomposition approach, as long as the assigned goal is solved (see  our prompts for three Execution experts in Appendix).
We expect that this flexibility will enable the experts to reason and use their "hands" in ways that align with their execution capabilities. Figure~\ref{fig:execution} illustrates an example of how an Execution Expert addresses an assigned short-term goal.

\paragraph{Assessment expert:}
The task of the Assessment expert is to evaluate the quality of the plan after receiving the result of the current short-term goal from the Execution Expert. Similar to the Planning expert, it does not execute tasks but focuses solely on reasoning, using an LLM specialized in assessment tasks. Its knowledge scope includes access to the table, the question, and the long-term plan (see our prompt for the Assessment in Appendix). After every \textit{k} short-term goals, the Assessment expert determines whether to generate an early answer if sufficient relevant information is available, or revise the plan if the results from the Execution experts fail to meet requirements or if the initial plan appears infeasible. Otherwise, the results from the Execution experts are automatically incorporated into the long-term plan. In essence, it takes a plan as input and outputs either a revised or an unchanged plan, or an answer.

\underline{Recommendation:} Our preliminary experiments show that frequent assessments, such as after each short-term goal, can facilitate early answers, conserve resources, and quickly address errors as they arise. However, frequent evaluations may also lead to challenges, such as overemphasizing short-term results at the expense of long-term objectives, unnecessary plan revisions (e.g., repeated short-term result validations), inaccurate premature answers, and increased resource costs. To mitigate these issues, \textit{k} should be tuned based on the data and the complexity of the question, balancing stability and efficiency.

\begin{table}[!t]
\centering
\begin{tabular}{l|c|c}
\hline
\textbf{Statistics} & \textbf{WikiTQ} & \textbf{TabFact} \\ \hline 
\# Questions & 4343 & 2024 \\
\# Number of Tables & 421  & 298 \\
\# Min/Max Rows & 6/518 & 5/49 \\
\# Min/Max Columns & 5/20 & 3/21 \\ \hline
\end{tabular}
\caption{Statistics of the WikiTableQuestions (WikiTQ) and TabFact test sets.}
\label{tab:dataset_statistic}
\end{table}

\section{Experiment Setup}
\paragraph{Dataset and Metric:} 
Following previous works \cite{wang2024chainoftable}, we conduct experiments on the benchmark datasets WikiTableQuestions---a question answering dataset over semi-structured tables \cite{pasupat2015compositional} and TabFact---a dataset for table-based fact verification \cite{Chen2020TabFact}. Table~\ref{tab:dataset_statistic} describes the statistics of their test sets. See a description of both datasets in the Appendix. 

We employ the official denotation accuracy \cite{pasupat2015compositional} for WikiTableQuestions  and the binary classification accuracy for TabFact.

\paragraph{Baselines:} 
We compare our method to recent strong table understanding methods, including \textbf{TEXT2SQL} \cite{rajkumar2022evaluating}, \textbf{CHAIN-OF-THOUGHT} \cite{wei2022chain}, 
\textbf{Dater} \cite{khot2023decomposed}, \textbf{StructGPT} \citep{jiang2023structgpt},  \textbf{BINDER} \cite{cheng2023binding}, \textbf{TabSQLify}  \cite{nahid-rafiei-2024-tabsqlify}, \textbf{CHAIN-OF-TABLE} \cite{wang2024chainoftable} and \textbf{DP\&PYAGENT} \cite{liu-etal-2024-rethinking}. CHAIN-OF-TABLE and DP\&PYAGENT are the state-of-the-art methods on TabFact and WikiTableQuestions, respectively.

\paragraph{Implementation Details:} 
We utilize \textit{LangGraph} to construct our proposed model, PLANTA, which is conceptualized as a graph.\footnote{https://langchain-ai.github.io/langgraph/} In this graph, the long-term plan represents the graph's state and each expert presents a node. Each expert is powered by an LLM with a distinct prompt, as detailed in Appendix . We mainly use {"GPT-3.5-turbo"} and {"GPT-4o-mini"} from OpenAI as the LLMs. The \textit{temperature} for LLMs is set to 0. The maximum number of iterations for a full turn of reasoning and  execution of predefined functions per expert is set to 2. The maximum number of short-term goals is set to 12.  The Assessment expert evaluates the long-term plan after completing $n-1$ steps of the plan where $n$ is the number of short-term goals in the plan. 

\section{Evaluation}

\subsection{Main Results}

Table~\ref{tab:main_result} reports the accuracy of our PLANTA and strong baselines on WikiTableQuestions (WikiTQ) and TabFact test sets. 


Recent state-of-the-art methods, including CHAIN-OF-TABLE and DP\&PYAGENT,  rely on chain-of-thought reasoning and self-consistency, demonstrating the effectiveness of these methods for table understanding. Both CHAIN-OF-TABLE and DP\&PYAGENT show notable improvements when upgrading their backbone LLM from GPT-3.5-turbo to GPT-4o-mini. For example, CHAIN-OF-TABLE improves from 59.9 to 70.4 on WikiTQ and 80.2 to 85.8 on TabFact. DP\&PYAGENT increases from 65.5 to 74.7 on WikiTQ and 80.0 to 89.9 on TabFact, highlighting the benefits of using a more powerful language model.

Our PLANTA outperforms all baselines on both test sets. With GPT-3.5-turbo, PLANTA scores 70.0 on WikiTQ and 82.0 on TabFact, outperforming DP\&PYAGENT (65.5 on WikiTQ, 80.0 on TabFact) and CHAIN-OF-TABLE (59.9 on WikiTQ, 80.2 on TabFact). When using GPT-4o-mini, PLANTA further improves to 75.7 on WikiTQ and 90.4 on TabFact, surpassing DP\&PYAGENT (74.7 on WikiTQ, 89.9 on TabFact) and CHAIN-OF-TABLE (70.4 on WikiTQ, 85.8 on TabFact).

Overall, PLANTA demonstrates state-of-the-art performance across different LLMs and datasets, providing clear evidence of the effectiveness of the proposed method for table understanding.

\begin{table}[t!]
\centering
\setlength{\tabcolsep}{0.3em}
\resizebox{7.5cm}{!}{
\begin{tabular}{l|cc}
\hline
\textbf{Method} & \textbf{WikiTQ} & \textbf{TabFact} \\ \hline
 & \multicolumn{2}{c}{GPT-3.5-turbo} \\
TEXT2SQL (2022) & 52.9 & 64.7  \\
CHAIN-OF-THOUGHT & 53.5 & 65.4  \\
BINDER (2023) & 56.7 & 79.2  \\
Dater (2023) & 52.8 & 78.0  \\
StructGPT (2023) & 48.4 & \_ \\
TabSQLify (2024) & 64.7 & 79.5  \\
CHAIN-OF-TABLE (2024) & 59.9 & \underline{80.2}  \\
 DP\&PYAGENT (2024) & \underline{65.5} & {80.0}  \\
Our PLANTA & \textbf{70.0} & \textbf{82.0}  \\ \hline
 & \multicolumn{2}{c}{GPT-4o-mini}  \\
 CHAIN-OF-TABLE & 70.4 &	85.8 \\
 DP\&PYAGENT  & \underline{74.7} & \underline{89.9} \\
Our PLANTA & \textbf{75.7} & \textbf{90.4} \\ \hline
\hline
\end{tabular}
}
\caption{
Performance results on the WikiTableQuestions (WikiTQ) and TabFact test sets. Rows 3 to 11 evaluate the table understanding capabilities of baseline methods and our PLANTA using GPT-3.5-turbo as the LLM. Results for previous methods are taken from their respective works, except for Dater, BINDER, and DP\&PYAGENT. Since original Dater and BINDER relied on the now-decommissioned OpenAI Codex LLM, we extract their results based on GPT-3.5-turbo, reported in the CHAIN-OF-TABLE paper \cite{wang2024chainoftable}. Furthermore, DP\&PYAGENT 
is tested only on a variant version of the original WikiTQ test set (i.e. not the same test set). Therefore, we run their official implementation (\url{https://github.com/Leolty/tablellm}) to report results on the original WikiTQ and the TabFact test sets with GPT-3.5-turbo. In rows 12-15, we run the official implementations of CHAIN-OF-TABLE (\url{https://github.com/google-research/chain-of-table}) and DP\&PYAGENT using GPT-4o-mini to provide results with a faster and more cost-efficient LLM. Note that \citet{wang2024chainoftable} also report results of CHAIN-OF-TABLE using "PaLM-2" with 340B parameters \citep{anil2023palm2technicalreport}. Since the PaLM-2 API has been decommissioned, we are unable to run PLANTA with "PaLM-2".
}
\label{tab:main_result}
\end{table}

\subsection{Ablation Study}
To investigate the impact of each proposed component of PLANTA, we evaluate our ablated variants on WikiTQ and TabFact.  Due to budget constraints, we evaluate the ablated variants on \textbf{1,000} randomly selected questions from each of the WikiTQ and TabFact test sets. Table~\ref{tab:ablation_study} presents the contribution of each proposed component to PLANTA's overall performance with GPT-4o-mini.

\textbf{W/o planning:} In this variant, long-term planning is excluded from PLANTA. Instead, the Planning expert relies solely on chain-of-thought (CoT) reasoning. In detail, the Planning expert is required to think step by step and generate a single request for Execution experts to handle. This process is repeated iteratively until a final answer is produced by the Assessment expert. As shown in Table~\ref{tab:ablation_study}, the exclusion of long-term planning significantly hurts PLANTA’s performance, with accuracy dropping from 76.5 to 69.0 on WikiTQ and from 90.0 to 74.0 on TabFact. Our internal analysis indicates that the sharper decline on TabFact is due to the nature of fact verification tasks, which typically require only a true/false response. This simplicity may cause the Assessment expert to prematurely decide on an answer without verifying supporting evidence. Meanwhile, WikiTQ questions, which involve more searching tasks, encourage the model to continue processing until the result is found, reducing premature mistakes.

\begin{table}[t!]
\centering
\resizebox{6.5cm}{!}{
\begin{tabular}{l|c|c}
\hline 
\textbf{Method} & \textbf{WikiTQ} & \textbf{TabFact} \\ \hline
PLANTA\textsubscript{GPT-4o-mini} & \textbf{76.5} & \textbf{90.0} \\ \hline
\ \ \ \ w/o planning & 69.0 & 74.0 \\
\ \ \ \ w/o search & 56.0 & 62.5 \\
\ \ \ \ w/o calculation & 71.5 & 81.5 \\
\ \ \ \ w/o comparison & 75.5 & 88.0 \\
\ \ \ \ w/o group experts & 74.4 & 88.0 \\
\ \ \ \ w/o assessment & 75.0 & 85.3  \\ \hline
\end{tabular}
}
\caption{The performance of the full-component PLANTA with GPT-4o-mini, along with the results of the ablation study.}
\label{tab:ablation_study}
\end{table}

\textbf{W/o search:} In this variant, the Search expert is excluded from PLANTA, and search tasks are instead handled by the Comparison and Calculation experts. This leads to a significant drop in accuracy, from 76.5 to 56.0 on WikiTQ and from 90.0 to 62.5 on TabFact, even though the brains of the Comparison and Calculation experts can still reason to perform searches. These results highlight that search is a critical task, and our design of the Search expert enables the brains to effectively utilize predefined functions, resulting in more accurate search performance compared to relying on reasoning alone.

\textbf{W/o calculation \& W/o comparison:} In these variants, the Comparison and Calculation experts are removed from PLANTA separately. Similar to the "W/o search" variant, these exclusions hurt PLANTA's accuracy. Specifically, removing the Calculation expert causes a sharper decline, with a 5\% drop on WikiTQ and 8.5\% on TabFact, compared to removing the Comparison expert, which results in a 1\% drop on WikiTQ and 2\% on TabFact, while the Search expert's brain still attempts reasoning to perform these tasks. These results highlight that LLMs' reasoning often struggles with comparison and even basic calculation.

\textbf{W/o group experts:} In this variant, all Execution experts are merged into a single unified expert responsible for handling search, calculation, and comparison tasks. Instead of using specialized prompts and predefined functions tailored to each expert’s specific task, the unified expert uses a general prompt and has access to all predefined functions. This consolidation results in a 2.1\% drop in accuracy on WikiTQ and a 2.0\% drop on TabFact. These results demonstrate that Execution experts benefit significantly from prompts and predefined functions designed specifically for their specialized tasks, highlighting the value of maintaining task-specific experts within PLANTA.

\textbf{W/o assessment:} In this variant, the Assessment expert is excluded from PLANTA. In details, all outputs from the Execution experts are automatically updated into the long-term plan, and the final answer is generated once all short-term goals are completed. Table \ref{tab:ablation_study} shows that removing the Assessment expert reduces PLANTA's accuracy by 1.5\% on WikiTQ and 4.7\% on TabFact. This discrepancy mainly arises from the need to revise the plan to handle code execution errors or situations in which one or more steps in the plan are infeasible, leading to repetitive iterations without returning valid results.


\begin{table*}[!t]
\centering
{\small
\begin{tabularx}{\linewidth}{lXc}
\hline
\textbf{Error Type} & \textbf{Description} & \textbf{\%} \\ \hline
Planning/Replanning & Errors related to incorrect relationships between steps, failure to handle exceptions within plan, and inability to detect execution errors when revising plan. & 37.7\% \\ 
Common sense & LLMs lack reasoning based on real-world knowledge. Example: When asked how many consecutive years 1990-1991 represent, LLMs answer \textit{"2"}, while the correct answer is 1. & 20.8\% \\ 
Lazy executor & Errors where experts rely solely on LLM reasoning, even when predefined functions could assist, leading to incorrect results. Example: LLMs miscalculate 3 + 3 + 1 = 6, but a tool could compute it correctly. & 11.7\% \\
Parameter errors & Errors caused by generating invalid parameters for predefined functions, such as wrong data types or conditions. & 11.3\% \\ 
Hallucination & The plan is executed correctly, but the conclusion is wrong. & 11.3\% \\ 
Acceptable answers & Unclear questions lead to answers that are technically correct but not aligned with the expected response. For example, when asked \textit{row listed before row 4}?, PLANTA lists rows 1 to 3, while the golden answer is row 3. & 5.7\%. \\ 
Missing "hands" & No predefined function is available to support the reasoning process & 1.9\% \\ \hline
\end{tabularx}
}
\caption{Error types in PLANTA\textsubscript{GPT-4o-mini} on the WikiTableQuestions test set. The total percentage does not add up to 100\% because some samples contain more than one error.}
\label{tab:error}
\end{table*}

\section{Analysis}



\subsection{Error analysis}
\label{subsec:error_analysis}

Table~\ref{tab:error} presents the types of errors observed in PLANTA. The most frequent errors are related to planning and common sense, stemming from the LLMs' lack of \textit{"real expert knowledge"}. As a result, they struggle to handle unpredictable data, such as \textit{"TBA"} for time or \textit{"note" columns} containing additional, contrasting information that alters the main context. This is consistent with the analysis in Subsection~\ref{subsec:planning_is_key}, where we demonstrate that improving the planning capabilities of the LLM leads to a substantial increase in accuracy. The Missing "hands" error, where no predefined function is available to assist reasoning, accounts for only 1.9\% of the cases, emphasizing the robustness of our design for predefined functions. However, 11.7\% of errors occur when the LLM mistakenly relies solely on reasoning instead of utilizing the available predefined functions to execute tasks accurately. In addition, 11.3\% of errors are caused by generating invalid parameters for predefined functions. Hallucinations remain an unavoidable issue with LLMs, accounting for 11.3\% of errors. LLMs can generate inaccurate final answers, even when accurate ones are explicitly provided in the final step. Despite its smaller percentage, acceptable answers reflect a need to handle vague questions.

\begin{figure}[!t]
    \centering
    \includegraphics[width=0.99\columnwidth]{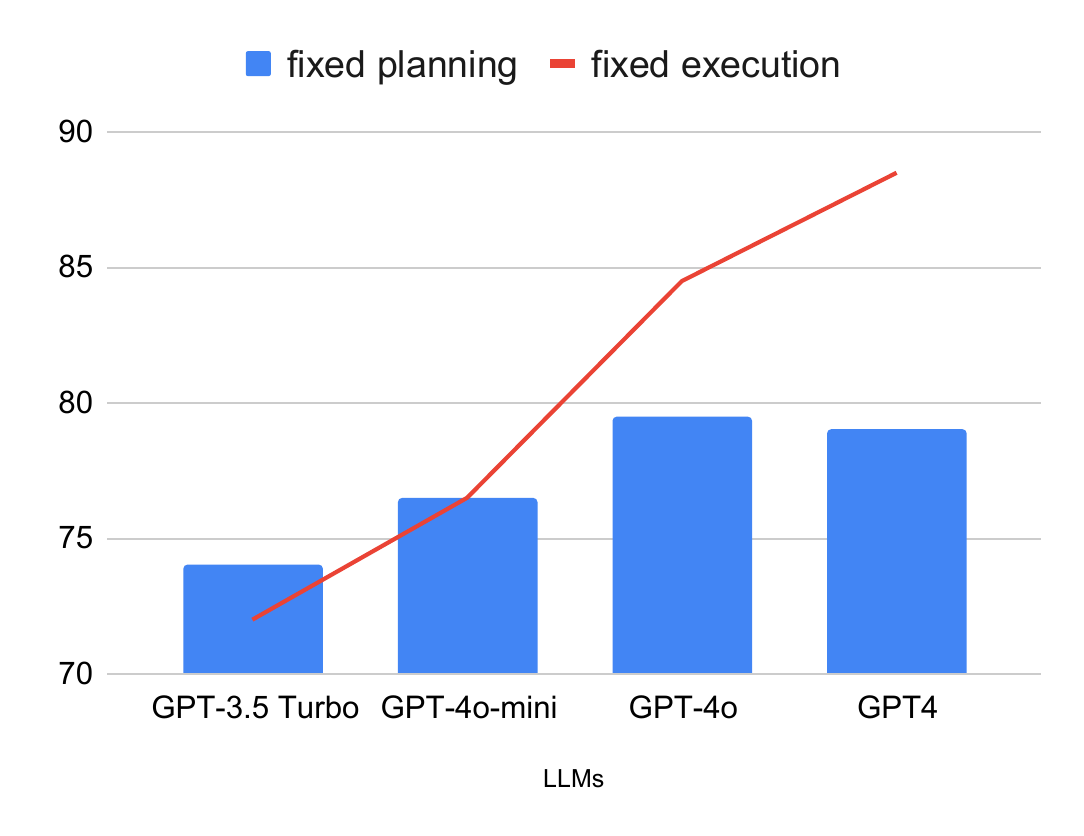}
    \caption{The impact of different LLMs on table understanding performance on the subset of 1000 WikiTableQuestions test questions used for the Ablation study. For "fixed planning", the LLM used for planning is set to GPT-4o-mini, while the LLM for execution tasks varies from GPT-3.5-turbo to GPT-4-mini to GPT-4o to GPT-4, increasing in reasoning capability. For "fixed execution", the LLM for execution tasks remains GPT-4o-mini, with the LLM for planning adjusted from GPT-3.5-turbo to GPT-4-mini to GPT-4o to GPT-4.}
    \label{fig:planning_is_key}
\end{figure}

\subsection{Improved LLMs Are Always the Key to Table Understanding?}
\label{subsec:planning_is_key}

As shown in Table~\ref{tab:main_result}, using LLMs with better reasoning capabilities notably improves table understanding performance. Here, we investigate whether improvements in LLMs always lead to significant performance gains. Our analysis focuses on two main aspects: [1] \underline{\textbf{Planning}}, which determines how PLANTA chooses the best approach to answer a question, including the task of planning by the Planning expert and the task of revised planning by the Assessment expert; and [2] \underline{\textbf{Execution}}, which involves performing the necessary tasks (such as search, comparison, and calculation) to find the relevant data within tables.

Figure~\ref{fig:planning_is_key} illustrates the impact of different LLMs on table understanding along these two aspects on the subset of 1000 WikiTableQuestions test questions used for the Ablation study. The results demonstrate that under the "fixed execution" setting, planning with better LLMs leads to a substantial improvement in accuracy for table understanding, with GPT-3.5-turbo achieving 72\% and GPT-4 increasing this to 88.5\%. In contrast, under the "fixed planning" setting, the improvement in execution tasks with better LLMs is far more limited, with accuracy rising from 74\% to 79\%. This contrast highlights the disproportionate influence of LLM reasoning on planning tasks compared to execution tasks. In other words, execution tasks appear to be less influenced by the model's reasoning power than planning tasks are, emphasizing the need for task-specific optimizations. Thus, by using powerful models for planning and more cost-effective models for execution, we can optimize both performance and resource efficiency.

\section{Conclusion}
We propose a novel method PLANTA to enhance table understanding by leveraging the long-term planning capabilities of LLMs. Our method focuses on two main goals: (1) enabling the execution of a long-term plan with tightly interconnected steps; (2) minimizing the inclusion of unnecessary details when solving short-term goals, thereby improving efficiency compared to CoT-based approaches. Experimental results show that PLANTA achieves new state-of-the-art performances on two benchmark datasets. Our PLANTA implementation is publicly available at: \texttt{https://github.com/nhungnt7/PLANTA}.

\section*{Limitations} 
Although our experiments have proven the effectiveness of our proposed method, there are still some limitations that can be improved in future work. While our approach encourages LLMs to engage in reasoning and solve tasks in a generalist manner, LLMs could benefit significantly from additional task-specific knowledge. For example, providing more targeted few-shot examples and explicitly including common exceptions could help the system handle rare or unpredictable scenarios better, as discussed in Subsection~\ref{subsec:error_analysis}.  Furthermore, future works can impose stricter constraints to encourage LLMs to use the "hands" of predefined functions effectively. This would minimize errors caused by LLMs attempting to rely solely on reasoning when predefined functions are better suited for the task, referred to as lazy executors in Subsection~\ref{subsec:error_analysis}.  

\section*{Acknowledgments}
This work was supported by Monash eResearch capabilities, including M3. 

This work was completed while Hoang Ngo and Dat Quoc Nguyen were at Movian AI, Vietnam.

\bibliography{custom}

\newpage 
\appendix

\section{Prompt}
\label{sec:prompt}
Table~\ref{tab:prompt} provides details of the custom-designed prompt for each component in the PLANTA system.
\begin{table*}[]
\centering
\begin{tabularx}{\linewidth}{XXX}
\textbf{Planning} & \textbf{Router} & \textbf{Search} \\ 
You are a Planning expert.  Your goal is to generate a plan to exclude a sequence of steps including SQL search (more detailed conditions in the requirements are better), calculation, and comparison based on the given table to get the answer to the question. For each step in the plan, dependencies on previous steps must be explicitly defined. 
Table: 
\textit{\{table\}}.
Question: 
\textit{\{question\}}.
& 
You are a task classification, your task is to classify the requirement type for the given task and route it to the appropriate expert. Please return the expert specialization based on the following guidance:
1. return 'search' if you need to search, conditional count the table for specific information. 
2. return 'compare' if you need to compare two or more pieces of information. 
3. return 'calculation' if you need to perform a calculation between numbers.
Your task:
\textit{\{short-term goal\}}.
&
You are a Search expert. You have been tasked to reason and generate an SQL query to extract and conditional count specific information (rows) from the table. We allow you to independently determine how to resolve the assigned goal, such as utilizing the Chain-of-Thought or question decomposition approach, as long as the goal is solved. You can use the tool to execute an SQL query generated based on the question and given table and return the result. You might know the answer without running any code, but you should still run the code to get the answer.
Given table:
\textit{\{table\}}. 
Your task:
\textit{\{shot-term goal\}} \\
&  &  \\
\textbf{Comparision} & \textbf{Calculation} & \textbf{Assessment} \\
You are a Comparison expert. You must use the tools provided to complete the assigned task.  We allow you to independently determine how to resolve the assigned goal, such as utilizing the Chain-of-Thought or question decomposition approach, as long as the goal is solved. You can use one tool multiple times and use many tools at one time in any order. You might know the answer without running any code, but you should still run the code to get the answer. 
Your tools include:
\textit{\{list of predefined functions\}}.
Your task:
\textit{\{shot-term goal\}}.
&
You are a Calculation expert. You must use the tools provided to complete the assigned task. We allow you to independently determine how to resolve the assigned goal, such as utilizing the Chain-of-Thought or question decomposition approach, as long as the goal is solved. You can use one tool multiple times and use many tools at one time in any order. You might know the answer without running any code, but you should still run the code to get the answer. 
Your tools include:
\textit{\{list of predefined functions\}}.
Your task:
\textit{\{shot-term goal\} }.
& 
You are an Assessment expert. Your goal is to answer the question if sufficient relevant information is available or revise the plan if the results from the Execution experts fail to meet requirements or if the initial plan appears infeasible.
Your original plan was this:
\textit{\{plan\}}.
You have currently done the follow steps with the following results at template (step, result):
\textit{\{past\_steps\}}
\end{tabularx}
\caption{Custom-designed prompts for each component in the PLANTA.}
\label{tab:prompt}
\end{table*}

\section{Dataset description}

\paragraph{WikiTableQuestions (WikiTQ):} A question answering dataset based on HTML tables, each with a minimum of 6 rows and 5 columns. The questions were not generated using predefined templates but were hand-crafted by users, resulting in significant linguistic diversity. These questions span various domains and require operations such as table lookup, aggregation, superlatives, arithmetic operations, joins, and unions.

\paragraph{TabFact:} A table-based binary fact verification dataset designed to determine whether a textual hypothesis is supported or refuted based on evidence provided in tables. The dataset presents a challenging task that requires both soft linguistic reasoning and hard symbolic reasoning. TabFact spans a wide range of operations, including aggregation, negation, superlatives, counting, comparative reasoning, and ordinal analysis.

\end{document}